\documentclass[10pt,conference]{IEEEtran}
\usepackage{amsmath}
\usepackage{graphicx}
\usepackage{booktabs}
\usepackage{float}
\usepackage[ruled,vlined]{algorithm2e}
\usepackage{listings}
\usepackage[table]{xcolor}
\usepackage{array}
\usepackage{makecell}
\usepackage{multirow}
\usepackage{cite}
\usepackage{url}
\usepackage{colortbl}
\definecolor{DeepRed}{RGB}{150,20,20}
\definecolor{LightRed}{RGB}{252,236,236}
\definecolor{TaskGreen}{RGB}{40,120,55}
\definecolor{CodeGray}{RGB}{70,70,70}
\definecolor{HeaderGray}{RGB}{245,245,245}
\definecolor{TableHeader}{RGB}{228,233,238}
\definecolor{TableStripe}{RGB}{244,247,249}
\definecolor{TableRule}{RGB}{72,78,86}
\definecolor{IconDeepSeek}{RGB}{45,105,170}
\definecolor{IconStarCoder}{RGB}{184,126,35}
\definecolor{IconQwen}{RGB}{42,132,126}
\definecolor{IconMistral}{RGB}{174,74,67}
\definecolor{IconCodeLlama}{RGB}{75,82,92}
\definecolor{IconSafeCoder}{RGB}{55,126,78}
\definecolor{IconLargeModel}{RGB}{105,84,145}
\definecolor{BestCell}{RGB}{225,225,225}   
\newcommand{\chg}[1]{%
  {\setlength{\fboxsep}{0.2pt}\colorbox{LightRed}{\textcolor{DeepRed}{#1}}}%
}
\newcommand{\modelbadge}[2]{%
  {\setlength{\fboxsep}{0.8pt}\raisebox{0.1ex}{%
    \colorbox{#1}{\textcolor{white}{\bfseries\scriptsize #2}}}}%
}
\newcommand{\modelhead}[3]{\modelbadge{#1}{#2}\,\textbf{#3}}
\lstdefinestyle{rollbackstyle}{
    language=Python,
    basicstyle=\ttfamily\scriptsize,
    keywordstyle=\color{DeepRed}\bfseries,
    commentstyle=\color{TaskGreen}\bfseries,
    stringstyle=\color{DeepRed},
    showstringspaces=false,
    keepspaces=true,
    columns=fullflexible,
    breaklines=true,
    breakatwhitespace=false,
    tabsize=4,
    xleftmargin=2pt,
    xrightmargin=2pt,
    frame=none,
    escapeinside={(*@}{@*)},
    aboveskip=\medskipamount,
    belowskip=\medskipamount
}
\lstdefinestyle{compactrollbackstyle}{
    style=rollbackstyle,
    basicstyle=\ttfamily\tiny,
    aboveskip=0pt,
    belowskip=0pt,
    xleftmargin=0pt,
    xrightmargin=0pt
}
\SetKwInput{KwInput}{Input}
\SetKwInput{KwOutput}{Output}
\SetKwComment{AlgoComment}{\texttt{// }}{}
\SetCommentSty{ttfamily}
\SetKw{KwReturn}{return}
\SetAlgoCaptionSeparator{:}
\begin{document}

\title{Do Uncertainty Signals Help?
A Systematic Study of Uncertainty-Aware Decoding with Rollback Mechanisms}

\author{
\IEEEauthorblockN{
Xianzong Wu$^{1}$,
Xiaohong Li$^{1}$,
Xinyang Liu$^{1}$,
Junjie Wang$^{1}$,
Qiang Hu$^{1}$,
Yuejun Guo$^{2}$,
Tianlin Li$^{3}$
}

\IEEEauthorblockA{
$^{1}$Tianjin University, China\\
$^{2}$Luxembourg Institute of Science and Technology, Luxembourg\\
$^{3}$Beihang University, China
}
}
\maketitle

\begin{abstract}

Prediction uncertainty is a widely adopted metric for quantifying model confidence, with downstream applications spanning model explanation, data selection, and prediction rollback. Despite its demonstrated utility, the potential of uncertainty quantification to enhance code generation in large language models (LLMs) remains largely underexplored, raising a critical question: \textit{to what extent can uncertainty serve as an effective signal for improving LLM-based code generation?}

To answer this question, we study uncertainty-aware rollback decoding, an inference-time strategy that uses uncertainty signals to identify unreliable generation regions and roll back to earlier valid prefixes without retraining the model. We evaluate this framework on seven code LLMs, five code generation benchmarks, and eight token-level uncertainty signals under a unified decoding setup.



Our results show that the complete rollback framework improves over equal-budget restart across the evaluated benchmarks and model settings, with gains of up to 0.26 in pass@1 and 0.35 in AvgTestPassRate on functional code generation benchmarks, and an absolute improvement of up to 6.4\% in Patch-Aligned Safe Rate on Dsec-Python.
Among the evaluated signals, information-theoretic measures such as token entropy and negative log-likelihood show the most favorable overall trend, frequently achieving the best or near-best results on standard benchmarks.
A component-controlled ablation further shows that feedback-guided rollback provides the main improvement, while uncertainty localization provides an additional gain when checking, budget, rollback, and branch decay are held fixed.
\end{abstract}

\section{Introduction}

Large language models (LLMs) have achieved strong performance on a wide range of code generation benchmarks, from function-level synthesis tasks to more realistic programming scenarios involving execution-based evaluation and complex instructions~\cite{chen2021codex,chen2022codet,liu2023evalplus,zhuo2025bigcodebench}. Despite this progress, code generation remains fundamentally fragile. Unlike natural language generation, code is governed by strict syntactic, semantic, and execution constraints, so a single early mistake can invalidate an entire program. Under standard left-to-right decoding, such errors are irreversibly incorporated into the generation prefix and may subsequently propagate through later statements, causing error accumulation and amplifying their downstream impact.

Prior work improves generated code through self-debugging, iterative refinement, execution guidance, and intermediate verification~\cite{chen2023selfdebug,madaan2023selfrefine,ni2023lever,zhang2023pgtd,le2024codechain}. Rollback-based decoding further enables failures to be corrected before a complete program is produced~\cite{jiang2025rocode}, while uncertainty estimation can identify unreliable generations~\cite{kuhn2023semantic,farquhar2024semanticentropy,nikitin2024kle}. However, post-generation repair acts only after errors have propagated, and uncertainty has mainly been studied as a final-answer reliability measure rather than a decoding-time control signal.

We therefore study whether uncertainty can guide rollback localization and regeneration. Our unified framework incrementally generates and checks code, rolls back after detected failures, and uses uncertainty to identify unreliable regions. We compare information-theoretic and sampling-based signals across multiple benchmarks and model scales under a fixed rollback mechanism.

Our study is organized around three research questions:
\begin{itemize}
    \item \textbf{RQ1: Does the complete rollback framework improve over equal-budget restart across datasets and models?}
    We answer this question by comparing the full framework with equal-budget restart on multiple benchmarks and model families. This comparison evaluates the complete framework rather than isolating uncertainty.

    \item \textbf{RQ2: What are the individual contributions of rollback, uncertainty, and branch decay?}
    Here, rollback denotes reverting to an earlier generation prefix after detecting failure, uncertainty denotes model-based signals for locating unreliable generation steps, and branch decay denotes a mechanism for reducing repeated exploration of unpromising branches. We answer this question through ablation experiments, rollback-localization diagnostics, and same-budget comparisons that isolate the effect of uncertainty-guided localization from additional generation budget.

    \item \textbf{RQ3: How does uncertainty behave across rollback-triggering error types?}
    We answer this question by analyzing the distribution of uncertainty scores across different categories of rollback-triggering errors.
\end{itemize}

Our results show that the complete rollback framework improves over equal-budget restart across the evaluated settings. The matched component ablation attributes the main improvement to feedback-guided rollback and an additional gain to uncertainty localization. Across the evaluated settings, information-theoretic signals generally provide more stable benefits than sampling-based alternatives.

In summary, this paper makes the following contributions:
\begin{itemize}
    \item We formulate uncertainty-aware rollback decoding as a decoding-time strategy for mitigating error accumulation in LLM-based code generation.
    \item We provide a controlled empirical comparison of information-theoretic and sampling-based uncertainty signals within a unified rollback framework.
    \item We present a multi-benchmark, multi-model evaluation of the complete rollback framework, together with a component-controlled uncertainty ablation and diagnostics of rollback localization, token cost, and same-budget alternatives.
\end{itemize}

\section{Problem Statement}

We formalize code generation as an autoregressive decoding process conditioned on a problem specification. 
Given an input prompt $x$ (e.g., a function signature together with a natural-language description), 
a pretrained language model generates an output token sequence 
$\mathbf{y}_{1:T} = (y_1, \ldots, y_T)$ of length $T$, 
where $y_t$ denotes the token at position $t$. 
We use $\mathbf{y}_{<t} = \mathbf{y}_{1:t-1}$ to denote the generated prefix before step $t$. 
The conditional distribution over the full sequence is factorized as
\[
p(\mathbf{y}_{1:T} \mid x) = \prod_{t=1}^{T} p(y_t \mid x, \mathbf{y}_{<t}).
\]

Standard left-to-right decoding cannot revise an erroneous prefix, so early code mistakes can propagate into later statements and execution failures~\cite{zhang2023pgtd,jiang2025rocode,olausson2024selfrepair,chen2021codex}. This is especially problematic for code, where one decision can make the remaining program syntactically invalid, semantically inconsistent, or non-executable.

Rollback-based decoding permits revision after a detected failure. Execution-based feedback has been used to assess generated code and guide inference-time revision or selection~\cite{ni2023lever,zhang2023pgtd,jiang2025rocode,li2022alphacode,le2022coderl,zheng2024opencodeinterpreter,bi2024cocogen}. In our framework, execution feedback determines \emph{whether} to reconsider the current prefix, while uncertainty refines \emph{where} rollback should occur~\cite{kuhn2023semantic,farquhar2024semanticentropy}.
Let $r < t$ denote a rollback position. 
When the current partial program corresponding to prefix $\mathbf{y}_{1:t}$ fails an execution check or a test case, 
the generated sequence is truncated back to the earlier prefix $\mathbf{y}_{<r} = \mathbf{y}_{1:r-1}$, 
and decoding resumes from position $r$. 
This changes generation from a strictly forward process into a search with selective revision. The key subproblem is selecting $r$: without a principled criterion, rollback remains a heuristic operation. We study uncertainty signals derived from the predictive distribution as a basis for this selection~\cite{kuhn2023semantic,farquhar2024semanticentropy,nikitin2024kle}. 
During decoding, each generated token $y_t$ is associated with an uncertainty score $u_t$, computed from the predictive distribution $p(\cdot \mid x, \mathbf{y}_{<t})$, for example using token entropy or negative log-likelihood. 
Tokens with high uncertainty indicate positions where the predictive distribution is diffuse, suggesting lower confidence and a higher likelihood of erroneous generation.

The central question is not only whether rollback helps, but whether uncertainty can indicate where rollback should occur. We therefore keep the rollback mechanism fixed while varying the uncertainty signal, enabling a controlled assessment against standard decoding and uncertainty-agnostic rollback.
\section{Uncertainty Signals}

Since rollback operates online, its uncertainty estimator must be efficiently computable and localized. We distinguish:
\emph{single-pass token-level signals}, which can be computed from a single forward pass at each decoding step, and \emph{multi-pass signals}, which require repeated forward passes to estimate predictive uncertainty more robustly.
This follows the standard separation between measures from one predictive distribution and those estimated through stochastic inference or committee-style disagreement~\cite{settles2009active,gal2016dropout,seung1992query}.

\subsection{Single-Pass Token-Level Signals}

Single-pass token-level signals are computed directly from $p(\cdot \mid x, y_{<t})$ without additional stochastic sampling.
We evaluate six such signals:

\begin{itemize}
    \item \textbf{Average Negative Log-Likelihood (Avg NLL)}: the average surprisal of generated tokens~\cite{shannon1948mathematical}.
    
    \item \textbf{Max Token Entropy}: the maximum token-level entropy within the current statement or span~\cite{shannon1948mathematical,settles2009active}.
    
    \item \textbf{Max Probability}: token confidence is converted to uncertainty as $1-\max_v p(v\mid x,y_{<t})$, and the maximum converted score within the statement is used~\cite{lewis1994sequential,settles2009active}.
    
    \item \textbf{Least Confidence}: the same token-level conversion, $1-\max_v p(v\mid x,y_{<t})$, is averaged over the statement. Thus, the two confidence measures share a common direction but differ in statement-level aggregation~\cite{lewis1994sequential,settles2009active}.
    
    \item \textbf{Margin of Confidence}: $1-(p_{(1)}-p_{(2)})$, where $p_{(1)}$ and $p_{(2)}$ are the two largest token probabilities; statement scores use the mean~\cite{settles2009active}.
    
    \item \textbf{Gini Impurity}: $1 - \sum_v p(v \mid x, y_{<t})^2$, capturing distributional dispersion~\cite{breiman1984cart}.
\end{itemize}

Avg NLL, Least Confidence, Margin of Confidence, and Gini Impurity use statement means; Max Token Entropy and converted Max Probability use statement maxima. Larger values consistently indicate higher uncertainty.

\subsection{Multi-Pass Signals}

Multi-pass signals require repeated stochastic forward passes. We consider two such signals:

\begin{itemize}
    \item \textbf{Predictive Entropy}: the entropy of the mean predictive distribution across stochastic samples~\cite{gal2016dropout,gal2016uncertainty}.
    
    \item \textbf{Variation Ratio}: $1 - f/N$, where $f$ is the modal-token frequency over $N$ samples~\cite{seung1992query,gal2016uncertainty}.
\end{itemize}

These signals require repeated stochastic inference and therefore cost more than single-pass signals~\cite{gal2016dropout}.

\subsection{Usage in Rollback Decoding}

Uncertainty complements rather than replaces execution feedback. Execution checks indicate \emph{whether} a generated prefix is problematic, while uncertainty helps locate \emph{where} to roll back. For single-pass signals, token scores are aggregated over the current statement span: Avg NLL uses mean token surprisal, Max Token Entropy uses maximum token entropy, and confidence-style signals are oriented so that larger values indicate lower confidence before statement-level ranking.

For multi-pass signals, token scores are computed from multiple stochastic forward passes and averaged within each statement before rollback is triggered~\cite{gal2016dropout,gal2016uncertainty}.
Table~\ref{tab:uq_signals_grouped} summarizes the uncertainty signals evaluated in this work and their grouping by computation type.

\begin{table}[ht]
\caption{Uncertainty signals considered in this work.}
\label{tab:uq_signals_grouped}
\resizebox{\columnwidth}{!}{%
\begin{tabular}{ll}
\arrayrulecolor{TableRule}
\toprule
\rowcolor{TableHeader}
\textbf{Computation Type} & \textbf{Signals} \\
\midrule
\rowcolor{TableStripe}
Single-pass token-level & \begin{tabular}[c]{@{}l@{}}Avg NLL, Max Token Entropy,\\ Max Probability, Least Confidence,\\ Margin of Confidence, Gini Impurity\end{tabular} \\
Multi-pass sampling-based & Predictive Entropy, Variation Ratio \\
\bottomrule
\end{tabular}%
}
\end{table}

All signals are integrated into the same rollback framework, so decoding, checking, and rollback conditions remain identical across estimators. Our goal is not to propose a new estimator, but to compare which signals best support rollback-based correction. We emphasize single-pass signals for online use, while multi-pass estimators serve as higher-cost references~\cite{gal2016dropout,settles2009active}.

\begin{figure*}[!t]
    \centering    
    \includegraphics[width=\textwidth]{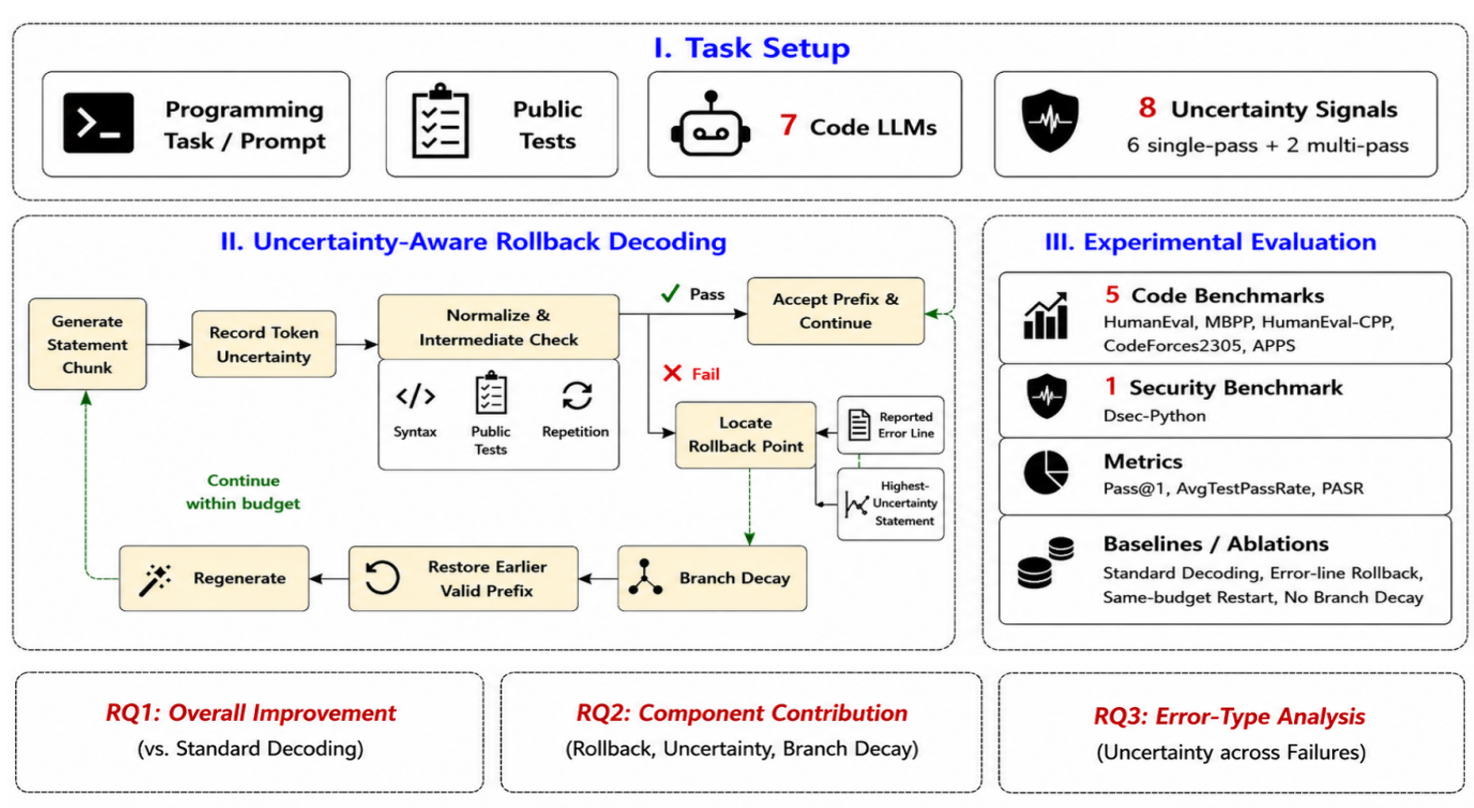}
    \caption{Overview of the study design. The task setup combines programming prompts, visible public tests, seven code LLMs, and eight uncertainty signals. During decoding, successful chunks extend the accepted prefix, whereas failed checks trigger feedback- or uncertainty-guided rollback, branch decay, prefix restoration, and regeneration. The evaluation addresses overall effectiveness, component contributions, and uncertainty across rollback-triggering error types.}
    \label{fig:framework}
\end{figure*}

\section{Uncertainty-Aware Decoding Framework}
Figure~\ref{fig:framework} illustrates the overall workflow of our framework.

Building on rollback-based code decoding~\cite{jiang2025rocode}, we instantiate a unified backbone that alternates incremental generation, lightweight checking, and rollback-based revision, while varying only the uncertainty signal. Given a task prompt and public tests, the decoder generates local chunks rather than the whole program at once. Execution feedback determines whether revision is needed, and uncertainty refines the rollback location. The goal is not a new rollback mechanism, but a controlled comparison under the same generation, checking, and regeneration protocol.

\paragraph{Incremental statement generation.}
As shown in Algorithm~\ref{alg:framework}, $y$ is the accepted prefix. At each iteration, the model extends it with a token-by-token, newline-bounded chunk $s$, which serves as a statement-like unit in practice. This enables validation after each local extension instead of waiting for the complete function.

Algorithm~\ref{alg:framework} uses the following notation. The candidate continuation is $c=y\oplus s$, where $\oplus$ denotes concatenation. $\operatorname{Norm}(c)$ makes an incomplete candidate executable using the normalization rules described below. $\operatorname{Check}$ returns either \textsc{Pass} or a diagnostic $e$ after syntax checking, public-test execution, and repetition detection. $\operatorname{Loc}(e)$ maps a valid syntax offset to its reported position, a repetition error to the beginning of the repeated line, and another execution error to the beginning of its reported line; it returns $\bot$ when that position is invalid. For each statement $S$ on the current path, $U(S)$ is the statement-level score of the selected uncertainty signal, using the aggregation defined in Section~III, and $\operatorname{start}(S)$ is its first token position. $\operatorname{Complete}(c,L)$ is true when generation emits the end-of-sequence marker or reaches the maximum length $L$.

The trie $\mathcal{G}$ stores generated tokens, their uncertainty values, and previously explored outgoing branches; $\operatorname{Path}(\mathcal{G})$ denotes the statements on its current candidate path. The counter $b$ includes all generated tokens, including discarded continuations. The pair $(\ell_{\mathrm{prev}},k)$ tracks the most recent feedback location and its consecutive failure count. Feedback localization is used for at most $K=2$ consecutive failures at one location; otherwise, rollback selects the highest-scoring statement on the current path. At the restored node, branch decay multiplies the stored sampling weight of the outgoing edge that begins the failed suffix by $\gamma=0.9$ before regeneration.

\begin{algorithm}[t]
\caption{Uncertainty-Aware Rollback Decoding}
\label{alg:framework}
\small
\DontPrintSemicolon
\KwInput{Prompt $x$, model $M$, public tests $\mathcal{T}$, maximum length $L$, token budget $B$}
\KwOutput{Completion $y$}
$y\leftarrow\emptyset$, $\mathcal{G}\leftarrow\operatorname{root}$, $b\leftarrow0$,
$(\ell_{\mathrm{prev}},k)\leftarrow(\bot,0)$\;
\AlgoComment{Step 1: Generate and record uncertainty}
\While{$b<B$ \textbf{and} $|y|<L$}{
    Generate a newline-bounded chunk $s$ from $M(\cdot\mid x,y)$\;
    Record every token and its uncertainty in $\mathcal{G}$\;
    $b\leftarrow b+|s|$, $c\leftarrow y\oplus s$\;
    \AlgoComment{Step 2: Validate the candidate}
    $z\leftarrow\operatorname{Norm}(c)$,
    $e\leftarrow\operatorname{Check}(z,\mathcal{T})$\;
    \eIf{$e=\mathrm{Pass}$}{
        $y\leftarrow c$, $(\ell_{\mathrm{prev}},k)\leftarrow(\bot,0)$\;
        \If{$\operatorname{Complete}(y,L)$}{
            \KwReturn $y$\;
        }
    }{
        \AlgoComment{Step 3: Select and apply rollback}
        $\ell\leftarrow\operatorname{Loc}(e)$\;
        \eIf{$\ell=\ell_{\mathrm{prev}}$}{
            $k\leftarrow k+1$\;
        }{
            $(\ell_{\mathrm{prev}},k)\leftarrow(\ell,1)$\;
        }
        \eIf{$\ell\neq\bot$ \textbf{and} $k\leq K$}{
            $r\leftarrow\ell$\;
        }{
            $r\leftarrow\operatorname{start}\!\left(\operatorname*{arg\,max}_{S\in\operatorname{Path}(\mathcal{G})}U(S)\right)$\;
        }
        Downweight by $\gamma$ the edge entering the failed suffix at $r$\;
        $y\leftarrow$ the prefix immediately preceding $r$\;
        Restore the current node of $\mathcal{G}$ to that prefix\;
    }
}
\KwReturn $y$\;
\end{algorithm}

During this process, the selected signal's token-level values are attached to nodes in $\mathcal{G}$ and aggregated into statement scores $U(S)$. These scores are consulted only when feedback localization is invalid or repeatedly fails.

\paragraph{Program assembly and feedback checking.}
The candidate partial program is normalized into executable form even when the function is incomplete. This includes completing indentation blocks and temporarily inserting \texttt{break} inside an open \texttt{while} block to avoid non-terminating intermediate execution. Lightweight analysis then performs syntax checking, public-test execution, and repetition detection. A passing chunk is merged into the accepted prefix; otherwise, the unreliable suffix is rolled back rather than discarding the whole program.

\paragraph{Rollback strategy.}
Unlike restart-based regeneration, rollback preserves the verified prefix. Reliable syntax offsets, repetition locations, and runtime error lines determine rollback directly. However, a runtime failure can surface after its underlying mistake, and shallow rollback can repeatedly revisit the same region. If the reported location is invalid or fails repeatedly, the decoder instead rolls back to the highest-uncertainty statement on the current trace, allowing it to bypass a larger suspicious region.

\paragraph{Generation trace and branch decay.}

The trie stores the accepted path together with previously explored continuations. On rollback, the decoder restores an earlier valid node and decays the outgoing edge that begins the failed suffix. This soft penalty reduces the probability of regenerating exactly the same continuation without permanently blocking it, preserving flexibility when the search later revisits the prefix.

\paragraph{Termination and budget control.}
The decoding loop is bounded by a token budget and a maximum generation length. When the model emits the end-of-sequence marker or reaches the maximum generation length, decoding terminates and returns the current completion. Hidden benchmark tests are applied only afterward for final evaluation and never affect online stopping or rollback.

Overall, the framework separates three complementary roles during decoding. Intermediate checking determines whether the current candidate remains executable, rollback preserves previously verified prefixes instead of restarting from scratch, and uncertainty is used only when explicit feedback cannot reliably localize the underlying error.

\section{Experimental Setup}

Unless otherwise stated, models are evaluated without retraining, and the rollback mechanism, checking procedure, and decoding loop remain fixed across conditions.

\subsection{Benchmarks}

We evaluate the framework on five code generation benchmarks and one security-oriented benchmark, covering function-level synthesis, multilingual execution, competition-style problem solving, and secure code repair~\cite{zheng2023surveyllmcode,cao2024javabench}.

\textbf{HumanEval}~\cite{chen2021codex} contains 164 function-level Python tasks with hidden unit tests. Intermediate checking uses executable public inputs when available; final evaluation follows the benchmark-standard pipeline.

\textbf{MBPP}~\cite{austin2021mbpp} provides shorter, structured Python synthesis tasks, complementing HumanEval with basic algorithmic and library problems.

\textbf{HumanEval-CPP}~\cite{zheng2023humanevalx} extends HumanEval-style evaluation to C++ and its stricter compilation constraints, where small structural errors can immediately invalidate the program~\cite{cao2024javabench}.

\textbf{CodeForces2305}~\cite{dong2024generalization} contains 90 recent competition problems emphasizing long-range reasoning, input/output handling, and algorithm design~\cite{li2022alphacode,zhuo2025bigcodebench}. The problems go beyond function-level synthesis, and their release dates reduce contamination risk for many evaluated models.

\textbf{APPS (Competition subset)}~\cite{hendrycks2021apps} provides difficult programming tasks with hidden tests and less constrained solution spaces, complementing the recent CodeForces problems with a broader standard benchmark.

\textbf{Dsec-Python}~\cite{he2024safecoder} evaluates whether generated security patches align with the intended secure edit region, using a patch-level metric rather than functional pass rate~\cite{bhuiyan2023secbenchjs,zhao2025securecodegen}.

\subsection{Decoding Setup}

The model generates a statement-level continuation up to a newline or local structural boundary, then immediately checks it before generating the next statement. This allows rollback to occur online rather than only after a full draft is produced~\cite{zhang2025actlcd,jiang2024selfplanning,fu2025firstprompt}.

Unless otherwise specified, decoding uses temperature $=0.0$, top-$k=50$, top-$p=1.0$, and one sample per task. 
The maximum generation length is set to 768 tokens, and the total token budget is limited to twice this length. 
For each task, error-line rollback is used for at most two consecutive failures at the same location before uncertainty-guided localization is activated. 
A path-decay factor of $0.9$ is applied during regeneration to reduce repeated exploration of previously failed continuations.

\subsection{Intermediate Checking and Rollback}

After each statement, benchmark-specific checks either accept the prefix or trigger rollback and regeneration from an earlier prefix.

We maintain a strict separation among lightweight checks, public tests, and hidden evaluation tests. Lightweight checks perform syntax validation, executable-prefix checks, and repetition detection during generation. Public tests are benchmark-provided visible examples used only for online checking and rollback decisions. Hidden tests are accessed only after generation terminates for final evaluation. No hidden-test result is exposed to decoding, rollback localization, branch decay, candidate selection, or early stopping.

Rollback localization follows a hybrid strategy. 
When explicit error information is available, such as syntax offsets, runtime error lines, or repetition-triggering regions, rollback is applied near the reported failure location. 
If this location is invalid or fails more than twice consecutively, the framework instead selects the statement containing the highest-uncertainty token in the current generation trace. 
This design separates two roles: execution or analysis feedback determines \emph{whether} the current prefix is problematic, while uncertainty helps determine \emph{where} the decoding trajectory should be revised~\cite{groninger2025changeguard}.

Path decay downweights, rather than blocks, previously failed continuations.

\subsection{Uncertainty Signals}

We evaluate the signals in Table~\ref{tab:uq_signals_grouped}. For each token position, multi-pass signals use $N=5$ stochastic forward passes with MC dropout and sampling temperature $1.0$. During uncertainty estimation, the model is temporarily switched to training mode to activate dropout, gradients are disabled, and the fixed per-task random seed is retained. The model is restored to evaluation mode for main-path generation. Predictive Entropy uses the mean probability distribution across the five passes; Variation Ratio uses their modal-token frequency~\cite{wang2025codeerrors}.
All uncertainty signals are used only for rollback localization; the surrounding decoding, testing, and rollback mechanism is kept unchanged across all settings.
Table~\ref{tab:uq_all} compares predictive effectiveness under the same rollback framework, but multi-pass signals require additional forward passes and the comparison is not compute-normalized.

\subsection{Baselines and Ablation Settings}

Our main RQ1 baseline is \textbf{equal-budget restart}. It uses the same prompt, model, sampling configuration, and 1,536-token total budget as the full framework. Each attempt is limited to 768 tokens; after a failed attempt, generation restarts from the original prompt within the remaining budget, without error feedback, rollback, path decay, or uncertainty localization.

Component ablations include rollback without uncertainty guidance and uncertainty scoring without rollback, separating the effects of rollback, localization, and branch decay.

We also conduct a same-budget comparison on HumanEval and CodeForces2305 to test whether the observed gains are simply caused by using more generation tokens. 
All compared strategies use CodeLlama-7B, temperature $=0.2$, and the same 1536-token budget; for each task, they share the initial seed defined as the base seed plus the task index. 
The first strategy, \textit{restart from scratch}, discards a failed attempt and regenerates a complete solution from the original prompt without error feedback, path decay, or sequence blocking. 
The second strategy, \textit{error-line rollback}, rolls back according to the reported error location without uncertainty-based refinement. Error-line rollback and the full framework use identical checking, regeneration, and branch-decay mechanisms; they differ only in whether uncertainty can refine the rollback location.
The third strategy is the full framework with uncertainty-guided rollback localization.
All strategies use public tests as the stopping signal.
Under the shared token budget, restart comparison controls for additional generation, while error-line rollback versus the full framework measures the incremental contribution of uncertainty-guided localization. The component ablation separately evaluates branch decay.

For Dsec-Python, equal-budget restart and the full framework use the same model, sampling configuration, and total token budget. The full framework permits at most three rollback recoveries within that budget, while restart may produce at most three candidates under the same budget. Both methods are counted as successful if any generated candidate satisfies PASR. PASR is computed only after generation and is not used for candidate selection, rollback, or stopping.

\subsection{Evaluation Metrics}

For HumanEval, MBPP, HumanEval-CPP, CodeForces2305, and APPS, we report \texttt{pass@1} and \textit{AvgTestPassRate}. For task $i$, let $T_i$ denote its hidden tests and $y_i$ the single generated program. Then
\[
\mathrm{AvgTestPassRate}=\frac{1}{|D|}\sum_{i=1}^{|D|}
\frac{1}{|T_i|}\sum_{j=1}^{|T_i|}\mathbf{1}[y_i\models t_{ij}].
\]
\texttt{pass@1} records whether $y_i$ passes all hidden tests for task $i$, whereas \textit{AvgTestPassRate} macro-averages the fraction of hidden tests passed per task. Thus, both metrics are compatible with one generated sample per task~\cite{chen2021codex,austin2021mbpp,hendrycks2021apps}.

For Dsec-Python, \textit{Patch-Aligned Safe Rate (PASR)} requires line-level IoU $\geq 0.5$ and normalized similarity $\geq 0.6$ in the modified region. A task succeeds if any of at most three candidates generated within the matched total token budget meets both criteria.

For the uncertainty-guided rollback diagnostics, we retain only events logged as \texttt{uncertainty\_fallback} or \texttt{uncertainty\_refinement}, excluding error-line and other feedback-only rollbacks. We report the number of uncertainty-selected events, the subset with a comparable reported error line, the within-3-line rate over that subset, and the fraction of uncertainty-selected events whose next lightweight check succeeds. The line-distance metric remains an operational proxy rather than manually annotated bug-localization ground truth.

For functional-error analysis, we use \texttt{AssertionError} as a proxy for functional failures, meaning programs that execute but fail test assertions; this should not be interpreted as a manually labeled semantic-error category.

\subsection{Implementation Details}

The framework uses PyTorch~\cite{paszke2019pytorch} and HuggingFace Transformers~\cite{wolf2020transformers}. The main experiments ran on four NVIDIA Tesla V100-SXM2 GPUs (16GB each), 28 Intel Xeon Gold 6132 CPU cores, and 754GB RAM, using Python 3.10, PyTorch 2.8.0, Transformers 4.40.0, and CUDA 12.7.

The rollback-diagnostic and same-budget experiments reported in Section~\ref{sec:diagnostics_same_budget} were run separately from the main multi-model evaluation using CodeLlama-7B and Max Token Entropy. 
We therefore interpret these results as within-environment comparisons among rollback strategies rather than as direct runtime comparisons with the main server experiments.
The diagnostic machine used an NVIDIA RTX 3090 (24GB), Intel Xeon Gold 6226R CPUs, 125GB RAM, and Ubuntu 20.04.6, with Python 3.11.9, PyTorch 2.3.0+cu121, and Transformers 4.40.1.

\section{Results}

This section evaluates overall effectiveness, uncertainty signals, model-scale effects, rollback diagnostics, component ablations, and error types. The main results are shown in Table~\ref{tab:rq1_overall}, Table~\ref{tab:uq_all}, Table~\ref{tab:rollback_diagnostics}, Table~\ref{tab:same_budget}, and Table~\ref{tab:rq2_ablation}, with complementary analyses in Figures~\ref{fig:ft_effect} and~\ref{fig:error_analysis}.

\subsection{RQ1: Does the complete rollback framework improve over equal-budget restart across datasets and models?}

Table~\ref{tab:rq1_overall} compares the complete rollback framework with equal-budget restart. Because the full framework jointly includes feedback, rollback, branch decay, and uncertainty localization, this comparison supports the complete framework rather than the isolated effect of uncertainty. Improvements appear on HumanEval, MBPP, and HumanEval-CPP. For example, on HumanEval, DeepSeek-7B improves from 0.32/0.49 to 0.52/0.69, Qwen2.5-7B from 0.36/0.56 to 0.55/0.77, and CodeLlama-34B from 0.48/0.68 to 0.61/0.81.

\begin{table*}[ht]
\centering
\scriptsize
\caption{RQ1 complete-framework results against equal-budget restart. The full framework uses Max Token Entropy. Both settings use a 1,536-token total budget and at most 768 tokens per attempt. Generation benchmarks report pass@1 / AvgTestPassRate; Dsec-Python reports PASR.}
\label{tab:rq1_overall}
\rowcolors{2}{TableStripe}{white}
\resizebox{\textwidth}{!}{
\begin{tabular}{lccccccc}
\arrayrulecolor{TableRule}
\toprule
\rowcolor{TableHeader}
\textbf{Dataset}
& \modelhead{IconDeepSeek}{DS}{DeepSeek-7B}
& \modelhead{IconStarCoder}{SC}{StarCoder2-7B}
& \modelhead{IconQwen}{Q}{Qwen2.5-7B}
& \modelhead{IconMistral}{M}{Mistral-7B}
& \modelhead{IconCodeLlama}{CL}{CodeLlama-7B}
& \modelhead{IconSafeCoder}{S}{SafeCoder-7B}
& \modelhead{IconLargeModel}{CL}{CodeLlama-34B} \\
\midrule
HumanEval & 0.52/0.69 & 0.45/0.68 & 0.55/0.77 & 0.40/0.61 & 0.39/0.61 & 0.36/0.59 & \cellcolor{BestCell}\textbf{0.61/0.81} \\
MBPP & 0.34/0.55 & 0.30/0.50 & 0.36/0.58 & 0.27/0.46 & 0.25/0.43 & 0.23/0.41 & \cellcolor{BestCell}\textbf{0.43/0.66} \\
HumanEval-CPP & 0.36/0.56 & 0.41/0.57 & 0.43/0.66 & 0.29/0.44 & 0.35/0.54 & 0.30/0.47 & \cellcolor{BestCell}\textbf{0.53/0.73} \\
CodeForces2305 & 0.09/0.28 & 0.08/0.26 & 0.10/0.32 & 0.07/0.24 & 0.06/0.20 & 0.05/0.18 & \cellcolor{BestCell}\textbf{0.15/0.40} \\
APPS-Competition & 0.16/0.26 & 0.13/0.21 & 0.15/0.24 & 0.14/0.25 & 0.10/0.18 & 0.07/0.10 & \cellcolor{BestCell}\textbf{0.23/0.34} \\
Dsec-Python & 0.56 & 0.51 & \cellcolor{BestCell}\textbf{0.57} & 0.48 & 0.55 & 0.53 & 0.54 \\
\midrule
HumanEval--Restart & 0.32/0.49 & 0.27/0.38 & 0.36/0.56 & 0.22/0.40 & 0.23/0.42 & 0.20/0.37 & 0.48/0.68 \\
MBPP--Restart & 0.22/0.35 & 0.19/0.32 & 0.23/0.37 & 0.17/0.29 & 0.16/0.27 & 0.14/0.26 & 0.34/0.54 \\
HumanEval-CPP--Restart & 0.18/0.30 & 0.20/0.27 & 0.17/0.31 & 0.15/0.27 & 0.23/0.33 & 0.16/0.24 & 0.43/0.61 \\
CodeForces2305--Restart & 0.05/0.15 & 0.04/0.14 & 0.06/0.18 & 0.04/0.13 & 0.03/0.11 & 0.03/0.10 & 0.11/0.30 \\
APPS-Competition--Restart & 0.13/0.24 & 0.08/0.17 & 0.14/0.19 & 0.09/0.15 & 0.08/0.18 & 0.05/0.16 & 0.16/0.34 \\
Dsec-Python--Restart & 0.50 & 0.47 & 0.52 & 0.45 & 0.49 & 0.47 & 0.50 \\
\bottomrule
\end{tabular}
}
\end{table*}

The trend remains visible on harder long-horizon benchmarks. On CodeForces2305, StarCoder2-7B from 0.04/0.14 to 0.08/0.26, and CodeLlama-34B from 0.11/0.30 to 0.15/0.40. 

For Dsec-Python, the complete framework also improves over equal-budget restart across all evaluated models in Table~\ref{tab:rq1_overall}, e.g., from 0.50 to 0.56 for DeepSeek-7B and from 0.52 to 0.57 for Qwen2.5-7B. However, the margins are smaller than those on execution-based benchmarks. A likely reason is that Patch-Aligned Safe Rate is a relatively coarse static metric, whose sensitivity to fine-grained decoding differences is limited.

\noindent\textbf{Finding 1.} The complete rollback framework improves over equal-budget restart across the evaluated benchmarks; this comparison does not isolate the contribution of uncertainty.

Table~\ref{tab:uq_all} compares the predictive effectiveness of uncertainty signals under the same rollback framework; it is not an equal-compute comparison because multi-pass signals require five forward passes per token position. Overall, information-theoretic signals, especially \textit{Max Token Entropy} and \textit{Avg NLL}, show the strongest trend on standard code-generation benchmarks. Max Token Entropy achieves the best or near-best results in many settings, including HumanEval, MBPP, HumanEval-CPP, and CodeForces2305. Avg NLL is also consistently competitive.

\begin{table*}[ht]
\centering
\scriptsize
\caption{Predictive-effectiveness comparison of uncertainty signals under the same rollback framework; results are not compute-normalized. Generation benchmarks report pass@1 / AvgTestPassRate; Dsec-Python reports PASR. Multi-pass signals use five stochastic forward passes. The best result for each dataset--model pair is shaded and boldfaced (ties included).}
\label{tab:uq_all}
\rowcolors{2}{TableStripe}{white}
\resizebox{\textwidth}{!}{
\begin{tabular}{llccccccc}
\arrayrulecolor{TableRule}
\toprule
\rowcolor{TableHeader}
\textbf{Dataset} & \textbf{Method}
& \modelhead{IconDeepSeek}{DS}{DeepSeek-7B}
& \modelhead{IconStarCoder}{SC}{StarCoder2-7B}
& \modelhead{IconQwen}{Q}{Qwen2.5-7B}
& \modelhead{IconMistral}{M}{Mistral-7B}
& \modelhead{IconCodeLlama}{CL}{CodeLlama-7B}
& \modelhead{IconSafeCoder}{S}{SafeCoder-7B}
& \modelhead{IconLargeModel}{CL}{CodeLlama-34B} \\
\midrule

\textbf{HumanEval}
& Avg NLL            & 0.49/0.67 & 0.45/0.62 & 0.54/0.72 & 0.38/0.56 & 0.42/0.59 & 0.34/0.55 & 0.60/0.77 \\
& Max Token Entropy  & \cellcolor{BestCell}\textbf{0.52/0.69} & 0.45/0.68 & \cellcolor{BestCell}\textbf{0.55/0.77} & 0.40/0.61 & 0.39/0.61 & 0.36/0.59 & \cellcolor{BestCell}\textbf{0.61/0.81} \\
& Max Probability    & 0.44/0.61 & 0.48/0.59 & 0.47/0.67 & 0.37/0.54 & 0.43/0.61 & 0.36/0.52 & 0.53/0.70 \\
& Least Confidence   & 0.46/0.63 & 0.43/0.64 & 0.51/0.65 & 0.42/0.58 & 0.40/0.57 & 0.31/0.53 & 0.52/0.72 \\
& Margin of Confidence & 0.47/0.65 & \cellcolor{BestCell}\textbf{0.50/0.68} & 0.49/0.69 & 0.36/0.55 & \cellcolor{BestCell}\textbf{0.44/0.58} & \cellcolor{BestCell}\textbf{0.37/0.51} & 0.56/0.71 \\
& Gini Impurity      & 0.51/0.62 & 0.41/0.66 & 0.55/0.68 & \cellcolor{BestCell}\textbf{0.43/0.60} & 0.39/0.56 & 0.35/0.58 & 0.54/0.69 \\
& Predictive Entropy & 0.43/0.66 & 0.49/0.61 & 0.48/0.73 & 0.39/0.61 & 0.36/0.53 & 0.33/0.59 & 0.57/0.78 \\
& Variation Ratio    & 0.45/0.60 & 0.46/0.67 & 0.50/0.64 & 0.35/0.57 & 0.37/0.54 & 0.30/0.50 & 0.51/0.74 \\
\midrule

\textbf{MBPP}
& Avg NLL            & 0.32/0.53 & 0.28/0.47 & 0.36/0.55 & 0.26/0.44 & 0.25/0.42 & 0.23/0.40 & 0.42/0.64 \\
& Max Token Entropy  & \cellcolor{BestCell}\textbf{0.34/0.55} & 0.30/0.50 & \cellcolor{BestCell}\textbf{0.36/0.58} & 0.27/0.46 & 0.25/0.43 & 0.23/0.41 & \cellcolor{BestCell}\textbf{0.43/0.66} \\
& Max Probability    & 0.29/0.49 & 0.32/0.44 & 0.30/0.52 & 0.25/0.41 & \cellcolor{BestCell}\textbf{0.28/0.46} & 0.25/0.38 & 0.37/0.60 \\
& Least Confidence   & 0.31/0.50 & 0.27/0.48 & 0.33/0.51 & 0.29/0.43 & 0.23/0.41 & 0.21/0.40 & 0.36/0.60 \\
& Margin of Confidence & 0.30/0.52 & \cellcolor{BestCell}\textbf{0.32/0.50} & 0.32/0.54 & 0.24/0.40 & 0.27/0.42 & \cellcolor{BestCell}\textbf{0.26/0.37} & 0.40/0.59 \\
& Gini Impurity      & 0.34/0.48 & 0.26/0.50 & 0.35/0.53 & \cellcolor{BestCell}\textbf{0.30/0.45} & 0.22/0.40 & 0.25/0.44 & 0.38/0.58 \\
& Predictive Entropy & 0.30/0.55 & 0.31/0.47 & 0.29/0.53 & 0.27/0.48 & 0.21/0.39 & 0.20/0.43 & \cellcolor{BestCell}\textbf{0.43/0.66} \\
& Variation Ratio    & 0.28/0.48 & 0.31/0.53 & 0.32/0.50 & 0.24/0.43 & 0.26/0.39 & 0.20/0.38 & 0.37/0.59 \\
\midrule

\textbf{HumanEval-CPP}
& Avg NLL            & 0.33/0.57 & 0.36/0.51 & 0.40/0.68 & 0.27/0.39 & 0.32/0.56 & 0.28/0.43 & 0.49/0.74 \\
& Max Token Entropy  & \cellcolor{BestCell}\textbf{0.36/0.56} & \cellcolor{BestCell}\textbf{0.41/0.57} & \cellcolor{BestCell}\textbf{0.43/0.66} & 0.29/0.44 & 0.35/0.54 & 0.30/0.47 & \cellcolor{BestCell}\textbf{0.53/0.73} \\
& Max Probability    & 0.29/0.46 & 0.38/0.54 & 0.35/0.61 & 0.26/0.37 & 0.37/0.50 & 0.32/0.41 & 0.45/0.70 \\
& Least Confidence   & 0.34/0.49 & 0.33/0.55 & 0.39/0.55 & 0.31/0.47 & 0.32/0.54 & 0.25/0.45 & 0.47/0.63 \\
& Margin of Confidence & 0.30/0.54 & 0.40/0.59 & 0.37/0.64 & 0.24/0.35 & \cellcolor{BestCell}\textbf{0.37/0.55} & \cellcolor{BestCell}\textbf{0.33/0.40} & 0.50/0.67 \\
& Gini Impurity      & 0.36/0.53 & 0.34/0.56 & 0.42/0.58 & \cellcolor{BestCell}\textbf{0.32/0.41} & 0.28/0.48 & 0.27/0.49 & 0.51/0.65 \\
& Predictive Entropy & 0.29/0.53 & 0.39/0.47 & 0.36/0.63 & 0.28/0.46 & 0.29/0.44 & 0.24/0.47 & 0.46/0.72 \\
& Variation Ratio    & 0.31/0.43 & 0.36/0.58 & 0.41/0.53 & 0.23/0.46 & 0.27/0.41 & 0.26/0.38 & 0.44/0.71 \\
\midrule

\textbf{CodeForces2305}
& Avg NLL            & 0.08/0.26 & 0.07/0.24 & 0.08/0.28 & 0.06/0.22 & 0.05/0.19 & 0.04/0.17 & 0.14/0.38 \\
& Max Token Entropy  & \cellcolor{BestCell}\textbf{0.09/0.28} & 0.08/0.26 & \cellcolor{BestCell}\textbf{0.10/0.32} & 0.07/0.24 & 0.06/0.20 & 0.05/0.18 & \cellcolor{BestCell}\textbf{0.15/0.40} \\
& Max Probability    & 0.07/0.23 & 0.08/0.23 & 0.07/0.26 & 0.06/0.19 & 0.06/0.20 & 0.05/0.16 & 0.12/0.35 \\
& Least Confidence   & 0.08/0.24 & 0.07/0.25 & 0.08/0.25 & 0.07/0.20 & 0.05/0.18 & 0.04/0.17 & 0.12/0.36 \\
& Margin of Confidence & 0.07/0.25 & \cellcolor{BestCell}\textbf{0.09/0.26} & 0.07/0.28 & 0.05/0.18 & \cellcolor{BestCell}\textbf{0.07/0.19} & \cellcolor{BestCell}\textbf{0.06/0.15} & 0.13/0.34 \\
& Gini Impurity      & 0.09/0.24 & 0.06/0.24 & 0.08/0.27 & \cellcolor{BestCell}\textbf{0.08/0.23} & 0.04/0.17 & 0.05/0.19 & 0.11/0.33 \\
& Predictive Entropy & 0.06/0.27 & 0.08/0.24 & 0.07/0.29 & 0.06/0.25 & 0.04/0.16 & 0.04/0.18 & 0.14/0.37 \\
& Variation Ratio    & 0.07/0.21 & 0.07/0.25 & 0.07/0.23 & 0.05/0.19 & 0.06/0.15 & 0.03/0.15 & 0.10/0.32 \\
\midrule

\textbf{APPS-Competition}
& Avg NLL            & 0.15/0.24 & 0.12/0.20 & 0.14/0.23 & 0.13/0.21 & 0.09/0.16 & 0.06/0.09 & 0.21/0.31 \\
& Max Token Entropy  & 0.16/0.26 & 0.13/0.21 & 0.15/0.24 & 0.14/0.25 & 0.10/0.18 & 0.07/0.10 & \cellcolor{BestCell}\textbf{0.23/0.34} \\
& Max Probability    & 0.11/0.19 & 0.15/0.22 & 0.10/0.18 & 0.12/0.17 & 0.11/0.19 & 0.08/0.11 & 0.18/0.27 \\
& Least Confidence   & 0.13/0.20 & 0.10/0.18 & 0.13/0.19 & 0.15/0.22 & 0.08/0.15 & 0.05/0.08 & 0.17/0.28 \\
& Margin of Confidence & 0.14/0.22 & \cellcolor{BestCell}\textbf{0.16/0.24} & 0.12/0.20 & 0.10/0.16 & \cellcolor{BestCell}\textbf{0.12/0.17} & \cellcolor{BestCell}\textbf{0.09/0.12} & 0.20/0.29 \\
& Gini Impurity      & \cellcolor{BestCell}\textbf{0.17/0.23} & 0.09/0.17 & \cellcolor{BestCell}\textbf{0.16/0.22} & \cellcolor{BestCell}\textbf{0.16/0.24} & 0.07/0.14 & 0.08/0.13 & 0.19/0.26 \\
& Predictive Entropy & 0.11/0.20 & 0.14/0.19 & 0.12/0.20 & 0.12/0.22 & 0.06/0.13 & 0.04/0.09 & 0.22/0.32 \\
& Variation Ratio    & 0.09/0.17 & 0.11/0.20 & 0.12/0.18 & 0.08/0.15 & 0.07/0.12 & 0.03/0.06 & 0.16/0.24 \\
\midrule

\textbf{Dsec-Python}
& Avg NLL            & 0.54 & 0.53 & 0.56 & 0.50 & 0.53 & 0.51 & 0.55 \\
& Max Token Entropy  & 0.56 & 0.51 & \cellcolor{BestCell}\textbf{0.57} & 0.48 & 0.55 & 0.53 & 0.54 \\
& Max Probability    & 0.51 & 0.56 & 0.52 & 0.51 & 0.55 & 0.50 & 0.52 \\
& Least Confidence   & 0.54 & 0.50 & 0.55 & 0.52 & 0.49 & 0.55 & 0.51 \\
& Margin of Confidence & 0.52 & \cellcolor{BestCell}\textbf{0.57} & 0.53 & 0.47 & \cellcolor{BestCell}\textbf{0.56} & 0.51 & 0.50 \\
& Gini Impurity      & \cellcolor{BestCell}\textbf{0.57} & 0.49 & 0.55 & \cellcolor{BestCell}\textbf{0.53} & 0.48 & \cellcolor{BestCell}\textbf{0.56} & 0.53 \\
& Predictive Entropy & 0.53 & 0.55 & 0.51 & 0.51 & 0.52 & 0.54 & \cellcolor{BestCell}\textbf{0.56} \\
& Variation Ratio    & 0.51 & 0.54 & 0.54 & 0.46 & 0.50 & 0.49 & 0.48 \\

\bottomrule
\end{tabular}
}
\end{table*}

At the same time, no single signal is universally optimal. In several cases, confidence-based or alternative signals remain competitive. For example, Gini Impurity performs strongly on APPS-Competition for DeepSeek-7B, and the best-performing signal on Dsec-Python varies by model. The differences among signals also become smaller on harder settings such as CodeForces2305, APPS-Competition, and Dsec-Python.

\noindent\textbf{Finding 2.} In the predictive-effectiveness comparison, Max Token Entropy and Avg NLL perform relatively better than other signals, but no signal consistently dominates; multi-pass results are not compute-normalized.

The gains in Table~\ref{tab:rq1_overall} are not limited to a single evaluated backbone. Improvements appear across multiple 7B models, including DeepSeek-7B, StarCoder2-7B, Qwen2.5-7B, Mistral-7B, CodeLlama-7B, and SafeCoder-7B, suggesting that the method is not tied to a specific architecture.

The framework also remains effective for stronger models. In Table~\ref{tab:rq1_overall}, CodeLlama-34B improves from 0.48/0.68 to 0.61/0.81 on HumanEval, from 0.34/0.54 to 0.43/0.66 on MBPP, and from 0.43/0.61 to 0.53/0.73 on HumanEval-CPP. This suggests that rollback remains useful not only for weaker models, but also for stronger generators in the evaluated settings.

\begin{figure}[ht]
    \centering    \includegraphics[width=\columnwidth,height=0.43\textheight,keepaspectratio]{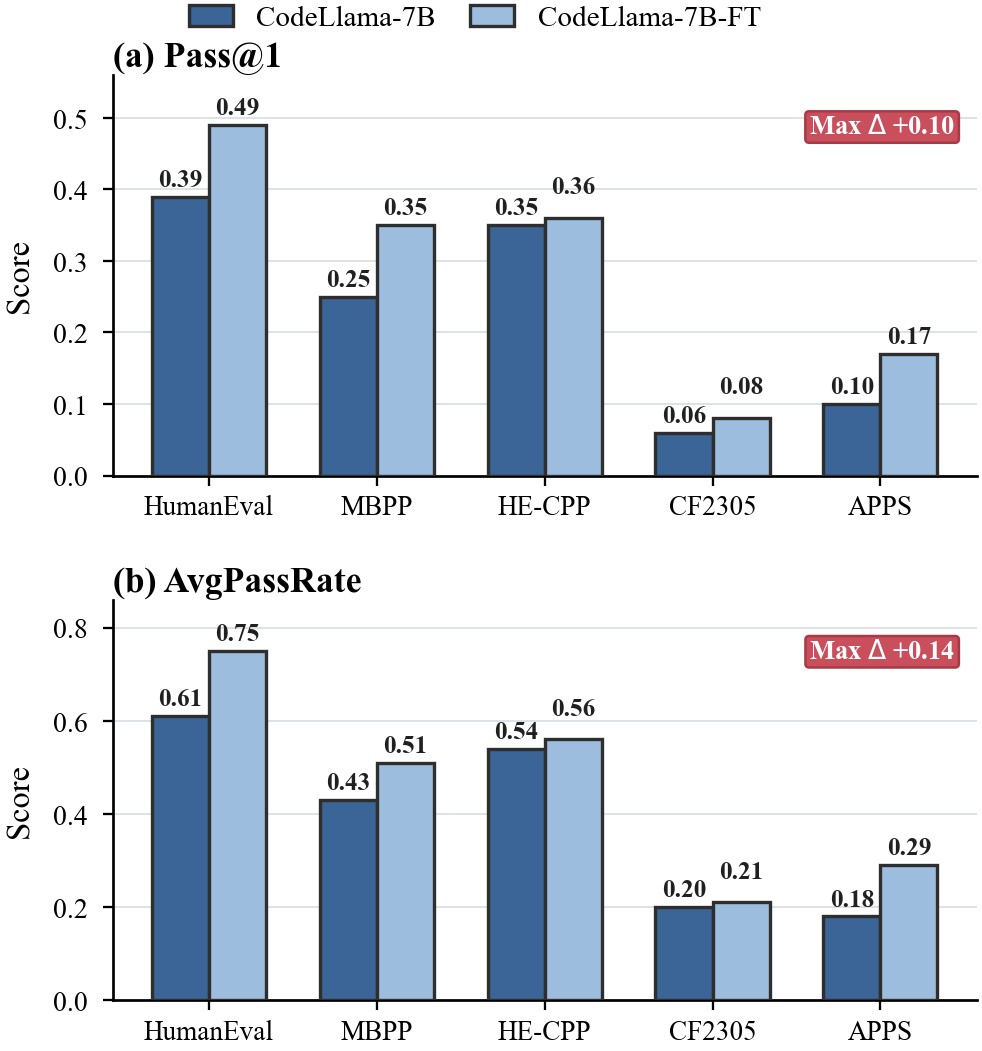}
    \caption{CodeLlama-7B and its fine-tuned variant under the same rollback configuration. Red labels show the maximum absolute gain in each panel.}
    \label{fig:ft_effect}
\end{figure}

We also fine-tune CodeLlama-7B on the APPS training split using 4-bit NF4 QLoRA for two epochs (learning rate $2\times10^{-4}$, rank 16, LoRA alpha 32, seed 42). Figure~\ref{fig:ft_effect} compares the merged final-epoch model with the base model under the same rollback configuration.

These improvements are consistently observed across models with different parameter scales and across both functional and security-oriented benchmarks. Although the absolute gains vary depending on task difficulty and model capability, the overall trend suggests that rollback-based decoding remains beneficial under diverse evaluation settings.

\noindent\textbf{Finding 3.} The proposed framework remains effective across the evaluated model scales, and the fine-tuned variant also improves over its base model.

\subsection{RQ2: What are the contributions of rollback, uncertainty, and branch decay?}
\label{sec:diagnostics_same_budget}

\subsubsection{Rollback Diagnostics and Controlled Comparison}

To examine uncertainty-guided decisions directly, we instrumented CodeLlama-7B with Max Token Entropy on HumanEval and on CodeForces2305 with its revised evaluation harness. The diagnostic run uses temperature $=0.0$ and seed $=0$. It records token entropy and, when the reported line is invalid or the same location fails more than twice, rolls back to the line containing the highest-entropy token on the current path; no fixed entropy threshold is used.
Table~\ref{tab:rollback_diagnostics} includes only \texttt{uncertainty\_fallback} and \texttt{uncertainty\_refinement} events. On HumanEval, 1,346 rollbacks are selected by entropy; 226 have a comparable reported error line, and 85.84\% of these selections fall within three lines. The next lightweight check succeeds after 77.93\% of uncertainty-selected rollbacks. On CodeForces2305, the corresponding counts are 951 and 417, with a 72.90\% within-3-line rate and a 70.14\% next-check pass rate.

\begin{table}[H]
\centering
\footnotesize
\caption{Diagnostics restricted to uncertainty-selected rollback events. UC Ev.: uncertainty-selected events; Comp.: events with a comparable reported error line; $\leq$3: rollback within three lines.}
\label{tab:rollback_diagnostics}
\setlength{\tabcolsep}{3.5pt}
\rowcolors{2}{TableStripe}{white}
\begin{tabular}{@{}lrrrr@{}}
\arrayrulecolor{TableRule}
\toprule
\rowcolor{TableHeader}
\textbf{Dataset} & \textbf{UC Ev.} & \textbf{Comp.} & \textbf{$\leq$3} & \textbf{Next Pass} \\
\midrule
HumanEval & 1,346 & 226 & 85.84\% & 77.93\% \\
CF2305 & 951 & 417 & 72.90\% & 70.14\% \\
\bottomrule
\end{tabular}
\end{table}

These measurements isolate behavior after entropy-based localization rather than mixing it with feedback-only rollback. They show that uncertainty-selected points are often close to the observed failure and frequently enable the next check to pass, although they do not by themselves constitute a counterfactual comparison against alternative localization rules.

The runs average 11.52 and 20.24 rollbacks per task on HumanEval and CodeForces2305, respectively. Uncertainty-score computation accounts for 3.99\% and 1.86\% of runtime in these diagnostic runs; these values do not represent end-to-end overhead relative to standard decoding.

We further compare the full framework against two alternatives under the same 1536-token budget. 
As shown in Table~\ref{tab:same_budget}, simply restarting from the original prompt after failure reaches 23.17\% pass rate on HumanEval, while rolling back directly to the reported error line reaches 32.32\%. 
The full framework reaches 41.46\% while using fewer average generated tokens than both alternatives. 
The same trend appears on CodeForces2305: restart and error-line rollback both reach 2.22\%, while the full framework reaches 4.44\%. 
Although the absolute pass rate on CodeForces2305 remains low due to task difficulty, the controlled comparison indicates that the gain is not explained merely by allowing more tokens or more regeneration attempts.

\begin{table}[H]
\centering
\footnotesize
\caption{Same-budget comparison under a 1536-token budget. Tok.: average generated tokens; RB: average rollback count (not applicable to restart).}
\label{tab:same_budget}
\setlength{\tabcolsep}{3pt}
\rowcolors{2}{TableStripe}{white}
\begin{tabular}{@{}llrrr@{}}
\arrayrulecolor{TableRule}
\toprule
\rowcolor{TableHeader}
\textbf{Dataset} & \textbf{Strategy} & \textbf{Pass} & \textbf{Tok.} & \textbf{RB} \\
\midrule
\textbf{HumanEval}
& Restart from scratch & 23.17\% & 1224.15 & -- \\
& Error-line rollback & 32.32\% & 935.34 & 72.15 \\
& Full framework & 41.46\% & 644.27 & 13.40 \\
\midrule
\textbf{CodeForces2305}
& Restart from scratch & 2.22\% & 1427.47 & -- \\
& Error-line rollback & 2.22\% & 1399.00 & 96.87 \\
& Full framework & 4.44\% & 1304.71 & 51.10 \\
\bottomrule
\end{tabular}
\end{table}

\paragraph{Functional-failure proxy.}
To distinguish assertion failures from parsing and runtime errors, we track \texttt{AssertionError}, which indicates that a program executes but fails a test assertion. In the temperature-$0.0$ diagnostic run, it accounts for 1,102/1,889 (58.34\%) rollback events and 69/164 (42.07\%) final samples on HumanEval, and 534/1,822 (29.31\%) rollback events and 72/90 (80.00\%) final samples on CodeForces2305. Event-level and final-sample statistics therefore use different denominators. In the separate temperature-$0.2$ same-budget run, final \texttt{AssertionError} proxy cases decrease from 110 to 76 on HumanEval and from 73 to 67 on CodeForces2305 when the full method replaces error-line rollback. This is a test-level proxy rather than manually annotated semantic-error ground truth.

\noindent\textbf{Finding 4.} Strictly filtered uncertainty-guided rollbacks frequently select locations near the reported failure and are followed by high next-check pass rates; uncertainty-score computation occupies a small fraction of diagnostic runtime, while same-budget comparisons show gains beyond extra generation budget and fewer final \texttt{AssertionError} proxy cases than error-line rollback.
An additional observation is that the framework often requires fewer generated tokens before converging to a successful solution. Instead of repeatedly regenerating complete programs, rollback revisits only suspicious regions while preserving verified prefixes. This behavior explains why the full framework can simultaneously achieve higher accuracy and lower average token consumption under the same overall decoding budget.
This result also suggests that rollback improves search efficiency instead of merely increasing exploration. Because validated prefixes are retained throughout decoding, the search process can concentrate on correcting suspicious regions rather than reconstructing already verified program fragments. Consequently, the available generation budget is allocated more effectively, allowing additional refinement attempts before the decoding budget is exhausted.

\subsubsection{Component Ablation}

Table~\ref{tab:rq2_ablation} reports the CodeLlama-7B ablation on HumanEval. Feedback-Only Detection improves only slightly over Vanilla Decoding (0.25 vs.\ 0.23 pass@1), while Feedback-Guided Rollback reaches 0.33 pass@1 and 0.53 AvgTestPassRate. The causal comparison for uncertainty is between Feedback-Guided Rollback and the Full Framework: both use feedback, rollback, branch decay, the same checks, and the same budget, and differ only in uncertainty localization. Adding uncertainty raises pass@1 from 0.33 to 0.39 and AvgTestPassRate from 0.53 to 0.61. This matched evidence is currently limited to CodeLlama-7B on HumanEval.

\begin{table}[H]
\centering
\footnotesize
\caption{CodeLlama-7B ablation results on HumanEval. Feedback-Guided Rollback and Full Framework share feedback, rollback, branch decay, checking, and budget; only the latter uses uncertainty localization. Comp. denotes compilation pass rate.}
\label{tab:rq2_ablation}
\setlength{\tabcolsep}{2.5pt}
\rowcolors{2}{TableStripe}{white}
\begin{tabular}{@{}lccc@{}}
\arrayrulecolor{TableRule}
\toprule
\rowcolor{TableHeader}
\textbf{Variant} & \textbf{pass@1} & \textbf{AvgTestPassRate} & \textbf{Comp.} \\
\midrule
Vanilla Decoding & 0.23 & 0.42 & 0.84 \\
Feedback-Only Detection & 0.25 & 0.44 & 0.86 \\
Feedback-Guided Rollback & 0.33 & 0.53 & 0.92 \\
Uncertainty-Guided Rollback & 0.27 & 0.46 & 0.88 \\
Full w/o Branch Decay & 0.37 & 0.59 & 0.94 \\
Full Framework & 0.39 & 0.61 & 0.96 \\
\bottomrule
\end{tabular}
\end{table}

\noindent\textbf{Finding 5.} Feedback-guided rollback provides the main improvement, while uncertainty provides an additional gain when checking, budget, rollback, and branch decay are controlled.

\subsection{RQ3: How does uncertainty behave across rollback-triggering error types?}

Table~\ref{tab:rollback_cases} complements the aggregate analysis with representative corrections across rollback-triggering error types.

\begin{table*}[!ht]
\centering
\scriptsize
\caption{Representative corrections across rollback-triggering error types on HumanEval. Highlighted tokens indicate the repaired fragments.}
\label{tab:rollback_cases}
\setlength{\tabcolsep}{2pt}
\renewcommand{\arraystretch}{0.82}
\begin{tabular}{p{0.15\textwidth} p{0.34\textwidth} p{0.34\textwidth}}
\arrayrulecolor{TableRule}
\toprule
\rowcolor{TableHeader}
\textbf{Error Type} & \textbf{Before Rollback} & \textbf{After Rollback} \\
\midrule

\textbf{NameError} &
\begin{minipage}[t]{\linewidth}\vspace*{-0.7\baselineskip}
\begin{lstlisting}[style=compactrollbackstyle]
# TASK 4
mean = sum(numbers) / len(numbers)
return sum(abs(x - (*@\chg{mean(numbers)}@*)) for x in numbers)
\end{lstlisting}
\end{minipage}
&
\begin{minipage}[t]{\linewidth}\vspace*{-0.7\baselineskip}
\begin{lstlisting}[style=compactrollbackstyle]
# TASK 4
mean = sum(numbers) / len(numbers)
return sum(abs(x - (*@\chg{mean}@*)) for x in numbers)
\end{lstlisting}
\end{minipage}
\\
\midrule

\textbf{IndexError} &
\begin{minipage}[t]{\linewidth}\vspace*{-0.7\baselineskip}
\begin{lstlisting}[style=compactrollbackstyle]
# TASK 33
l.insert(i, sorted(l[i:i+3]))
l.pop(i+1)
l.pop(i+1)
(*@\chg{l.pop(i+1)}@*)
\end{lstlisting}
\end{minipage}
&
\begin{minipage}[t]{\linewidth}\vspace*{-0.7\baselineskip}
\begin{lstlisting}[style=compactrollbackstyle]
# TASK 33
(*@\chg{l[i:i+3] = sorted(l[i:i+3])}@*)
\end{lstlisting}
\end{minipage}
\\
\midrule

\textbf{RepeatPatternError} &
\begin{minipage}[t]{\linewidth}\vspace*{-0.7\baselineskip}
\begin{lstlisting}[style=compactrollbackstyle]
# TASK 1
if paren_string == '()':
    return ['()']
if paren_string == '()()':
    return ['()', '()']
(*@\chg{\texttt{if paren\_string == '()()()':}}@*)
\end{lstlisting}
\end{minipage}
&
\begin{minipage}[t]{\linewidth}\vspace*{-0.7\baselineskip}
\begin{lstlisting}[style=compactrollbackstyle]
# TASK 1
groups, cur, bal = [], [], 0
for ch in paren_string:
    bal += 1 if ch == '(' else -1
    cur.append(ch)
    if bal == 0:
        groups.append(''.join(cur)); cur = []
\end{lstlisting}
\end{minipage}
\\
\midrule

\textbf{SyntaxError} &
\begin{minipage}[t]{\linewidth}\vspace*{-0.7\baselineskip}
\begin{lstlisting}[style=compactrollbackstyle]
# TASK 6
for ch in paren_string:
    if ch == '(':
        depth += 1
return (*@\chg{[}@*)
\end{lstlisting}
\end{minipage}
&
\begin{minipage}[t]{\linewidth}\vspace*{-0.7\baselineskip}
\begin{lstlisting}[style=compactrollbackstyle]
# TASK 6
for ch in paren_string:
    if ch == '(':
        depth += 1
return (*@\chg{depth}@*)
\end{lstlisting}
\end{minipage}
\\

\bottomrule
\end{tabular}
\end{table*}

Table~\ref{tab:rollback_cases} illustrates that rollback supports both localized repairs and broader structural revisions. NameError and SyntaxError are corrected through small fragment replacements, whereas IndexError and RepeatPatternError require rewriting a larger statement or control-flow block, showing why rollback depth should adapt to the observed failure pattern.

\begin{figure}[ht]
    \centering
    \includegraphics[width=\columnwidth]{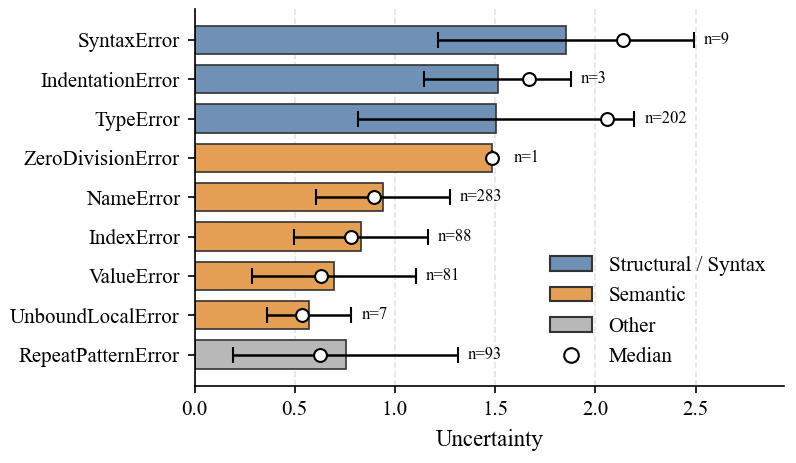}
    \caption{Rollback-triggering error types and uncertainty. Bars show event-level means, white circles show medians, and error bars denote one standard deviation across rollback events. Error bars are omitted for categories with $n=1$.}
    \label{fig:error_analysis}
\end{figure}

Figure~\ref{fig:error_analysis} shows the distribution of rollback-triggering error types and their associated uncertainty values. NameError ($n=283$) and TypeError ($n=202$) are among the most frequent error categories, followed by RepeatPatternError, IndexError, and ValueError.

At the same time, the highest uncertainty values are associated with more structurally unstable errors. SyntaxError has the highest mean uncertainty (1.85), while IndentationError (1.51) and TypeError (1.51) are also clearly above the overall average. In contrast, more common local semantic errors such as NameError (0.94), IndexError (0.83), and ValueError (0.70) show lower uncertainty. This suggests that uncertainty is not simply tracking how often an error occurs; instead, it better reflects how destabilizing that error is to the ongoing generation process.

The observed uncertainty distribution also suggests that uncertainty is associated with structural instability rather than merely reflecting error frequency. This observation is consistent with the intuition that rollback should prioritize regions where the decoding trajectory becomes unstable instead of simply revisiting the most recently generated tokens.

This pattern supports the role of uncertainty in our framework: it is more useful as a cue for rollback localization than as a standalone predictor of final correctness. 
We note that categories with fewer than five events are reported descriptively and should be interpreted cautiously; the error bars summarize dispersion and are not confidence intervals.

\noindent\textbf{Finding 6.} Rollback-triggering uncertainty tends to be higher for structurally unstable errors than for common local semantic mistakes, supporting its use as a rollback-localization cue.

\section{Threats to Validity}

\textbf{Statistical validity.}
Due to computational cost, our experiments use a single-run evaluation setting and do not include repeated trials, confidence intervals, or formal significance tests. Therefore, the reported improvements should be interpreted as empirical trends on the evaluated benchmarks rather than statistically conclusive effect estimates, especially when the gaps between uncertainty signals are small~\cite{liu2023evalplus,zhuo2025bigcodebench,dong2024generalization}.

\textbf{Construct validity.}
Our evaluation uses functional correctness metrics (e.g., pass@1 and AvgTestPassRate) for code-generation benchmarks and Patch-Aligned Safe Rate for Dsec-Python. While practical, these metrics do not fully capture all aspects of generation quality. In particular, the Dsec-Python metric relies on patch alignment and lightweight static validation, so its sensitivity to subtle security improvements is limited compared with end-to-end exploit-level verification.
Similarly, our rollback-line distance and \texttt{AssertionError} analyses are operational proxies rather than manually annotated ground truth for bug localization or semantic functional-error categories.

\textbf{External validity.}
Although we evaluate multiple representative open-source code LLMs, including several 7B-scale models and one 34B model, the study still covers only a limited portion of the model space. In addition, most benchmarks are Python-centered. Therefore, the findings should be understood as evidence on the evaluated settings rather than a universal conclusion.

\textbf{Methodological validity.}
Our conclusions are drawn under a fixed rollback-based decoding framework with a specific intermediate-checking and branch-decay design. Therefore, the observed effectiveness of uncertainty signals should be interpreted in this context. Different rollback strategies, checking mechanisms, or search policies may lead to different relative behaviors of uncertainty estimators. In addition, lightweight intermediate checking may not capture all failure modes equally well.
The diagnostic and same-budget experiments were conducted on a separate machine from the main multi-model evaluation, so their runtime numbers should be interpreted within that diagnostic setting rather than directly compared with the main experimental server.

Future work may further investigate adaptive rollback policies, alternative intermediate verification strategies, and larger proprietary code models. We also expect stronger execution feedback to further improve uncertainty-guided localization in more realistic programming environments.

\section{Conclusion}

This paper studied whether uncertainty signals can improve rollback-based decoding for code generation. The complete rollback framework improves over equal-budget restart across the evaluated models and benchmarks, with gains of up to 0.26 in pass@1 and 0.35 in AvgTestPassRate on functional benchmarks and up to 6.4\% in Patch-Aligned Safe Rate on Dsec-Python. These comparisons evaluate the complete framework rather than uncertainty alone. In the matched component ablation, feedback-guided rollback provides the main improvement, while uncertainty adds a further gain when checking, budget, rollback, and branch decay are controlled. Diagnostics restricted to entropy-selected events show high within-3-line and next-check pass rates; uncertainty-score computation accounts for a small fraction of diagnostic runtime but does not measure end-to-end overhead relative to standard decoding. Information-theoretic signals show the most stable predictive effectiveness, although multi-pass comparisons are not compute-normalized and no signal is universally best.

\section*{Data Availability Statement}

The replication package, including the source code and evaluation scripts, is available through an anonymous repository: https://anonymous.4open.science/r/ICSE2027-257C

\bibliographystyle{IEEEtran}
\bibliography{refs}

\end{document}